\documentclass{CUP-JNL-NLP}

\usepackage{amsmath}
\usepackage{graphicx}
\usepackage{natbib}
\ifpdf%
\usepackage{epstopdf}%
\else%
\fi
\usepackage{multirow}

\begin{document}
\label{firstpage}

\lefttitle{\LaTeX\ Supplement}
\righttitle{Natural Language Processing}

\papertitle{Article}

\jnlPage{\pageref{firstpage}}{\pageref{lastpage}}

\title{Assisting humans in complex comparisons: automated information comparison at scale}

\begin{authgrp}
\author{Truman Yuen}
\affiliation{Centre for Management of Technology and Entrepreneurship, University of Toronto, Toronto, M5S 1E1, Ontario, Canada}

\author{Graham A. Watt}
\affiliation{Royal Bank of Canada, Toronto, Ontario, Canada}

\author{Yuri Lawryshyn}
\affiliation{Centre for Management of Technology and Entrepreneurship, University of Toronto, Toronto, M5S 1E1, Ontario, Canada}
\end{authgrp}

% \history{(Received xx xxx xxx; revised xx xxx xxx; accepted xx xxx xxx)}
%\received{20 March 1995; revised 30 September 1998}

\begin{abstract}
Generative Large Language Models enable efficient analytics across knowledge domains, rivalling human experts in information comparison. However, applying LLMs for information comparison faces scalability challenges due to difficulties in maintaining information across large contexts under model token constraints. State-of-the-art (SOTA) text comparison solutions are limited in applicability due to their dependency on domain-specific training data, computational requirements and data scalability. Addressing these barriers enables rapid implementation and adoption of applied text comparison using LLMs across domains. In this study, we propose our Abstractive Summarization \& Criteria-driven Comparison Endpoint (ASC$^2$End) system to automate information comparison at scale. ASC$^2$End combines abstractive summarization and retrieval augmented generation (RAG), creating a novel pre-retrieval RAG process to overcome token limitations and retain relevant information during model inference and uses data-handling strategies with advanced prompting techniques to establish a new paradigm in pre-retrieval RAG for information comparison tasks, eliminating the need for extensive domain-specific training. ASC$^2$End incorporates Semantic Text Similarity comparisons to generate evidence-supported analyses. System performance was evaluated using precision metrics and human survey responses. ASC$^2$End showed significant overall accuracy (94\%), with superior efficiency and comparable runtimes versus the baseline. ASC$^2$End is a novel system and tool that enables accurate, automated information comparison at scale for applications across financial services and other knowledge domains.
\end{abstract}

\maketitle

\section{Introduction}

The applications of generative Large Language Models (LLMs) across knowledge domains, including financial services, are rapidly expanding due to their proven ability to perform multiple tasks such as abstractive summarization \citep{S.Liu_2023}, simple QA (Question-Answering) \citep{Trivedi_2022}, multiple choice QA \citep{Trivedi_2022}, financial sentiment analysis \citep{Fei_2023}, Retrieval Augmented Generation (RAG) \citep{Pan_2022}, among others \citep{Lu_2021}. These open-domain tasks have recently been popularized in specific applications, enabling time efficiencies in the analytics required for informed decision-making.

This study defines generative LLMs as exclusively autoregressive models. Recently, generative LLMs have proven to be effective in executing abstractive summarization \citep{Rath_2023}, outperforming existing state-of-the-art (SOTA) models. \cite{S.Liu_2023} presented the concept of abstractive summarization through semantic splitting instead of the traditional token length split, which may present better-contextualized summaries. Similarly, \cite{dixit_2023} proposes training LLMs on labelled datasets to increase information retention during inference.

Fine-tuning generative LLMs increases a model's overall performance in a specific domain \citep{Brown_2020}. However, prompting strategies leverage the performance capabilities of naive generative LLMs without relying on large, domain-specific training data sets, with some minor trade-offs in performance \citep{Navarro_2022}\citep{Howell_2023}. Zero-shot \citep{Kojima_2022} and few-shot \citep{Wei_2022} prompting techniques can be leveraged to direct answer structures and elicit human-like reasoning for response generation. A recent study demonstrated the ability of naive generative LLMs to perform Named Entity Recognition (NER) and Relation Extraction (RE) \citep{Li_2023} for tasks that reflect human-like reasoning. Chain-of-thought reasoning can also be elicited in zero-shot prompting strategies using phrases such as “let’s think step-by-step” \citep{Kojima_2022} to reflect human-like reasoning. Additionally, altering the available input context and the order in which information is presented in a prompt \citep{Lu_2021} can affect the final output response structure.

Semantic textual similarity (STS) is the main driving factor in effective information comparison \citep{Majumder_2016}. STS is measured through many different techniques that can help in text classification and topic extraction \citep{Slimani_2013} without the limitations of lexical similarity. Semantic similarity is achievable with generative LLMs as demonstrated by \citet{Gatto_2023}. They achieved success in assessing semantic similarity by testing various prompting strategies on generative LLMs. Moreover, using retrieval augmented generation (RAG) provides additional context based on a user query such that an LLM can generate an in-domain output response \citep{Ram_2023} \citep{Trivedi_2022}. RAG applies STS to find the top-\textit{k} passages most similar to the user query and enhances the information in its output response. \cite{gao_2024} outlines the current advances in RAG techniques and the categorization of similar technologies. Modular components of RAG pipelines are classified as advanced RAG techniques where pre and post-retrieval data processing is applied to better augment the output of RAG.

Recent research has demonstrated that decision-making and data analysis can substantially benefit from text analyses made available by STS through generative LLMs. For example, research in the medical domain has developed a framework to screen abstracts of scientific papers to be used for review papers \citep{Guo_2023}. \citet{Guo_2023}'s work used generative LLMs to compare scientific abstracts with user-defined criteria and sorted the abstracts based on the eligibility generated by the model. Their work provided insights into using generative LLMs to perform information comparisons against user-defined criteria such that the model could make binary categorical decisions. Similar research was also conducted in the financial domain, where a corporate sustainability report was compared to a sustainability guideline document in their framework, chatReport \citep{Ni_2023}. chatReport supported different QA tasks regarding the information in the sustainability report, powered by RAG. Their framework handles individual reports and generates responses for single-use applications.

However, applying LLMs for information comparison is currently non-trivial at scale due to token limitations imposed on many LLMs \citep{sun2023}. Minimizing information loss and prompting under token limits are ongoing issues that must be addressed to expand system functionality \citep{jaiswal2023}. Models with longer token limits are prone to losing information from the input context, for example, due to limitations in relevant information retrieval, from the middle of long input contexts \citep{N.Liu_2023}. \citet{N.Liu_2023} found that naive, untrained models with no input context performed better on the same QA task when compared to the models provided with a lengthy input context. Given these challenges, new strategies must be explored to enhance effective and efficient information retrieval while overcoming token limitations.

This study aims to develop an advanced LLM framework that improves text comparison accuracy and scalability by addressing token limitations and contextual information loss while minimizing computational resources. We propose ASC$^2$End (\textbf{A}bstractive \textbf{S}ummary \& \textbf{C}riteria-driven \textbf{C}omparison \textbf{End}point), a novel system enabling accurate, automated information comparison at scale for applications across financial services and other knowledge domains. Through abstractive summarization, RAG and prompt engineering, ASC$^2$End introduces a pre-retrieval RAG workflow that enables efficient, large-scale information comparison across knowledge domains without extensive domain-specific knowledge. Existing research in pre-retrieval RAG processes is limited by the complexity and volume of text provided to its process \citep{zheng_2024}\citep{ma_2023} and abstraction research exhibits dependencies in extensive model finetuning \citep{dixit_2023}. ASC$^2$End robustly handles complex RAG tasks and eliminates the need for model fine-tuning. We demonstrate the applicability of our system to the challenge of identification and evaluation of financial transactions against complex, user-defined sustainable finance criteria. We quantify our results by evaluating the performance of prominent LLMs using ROUGE and survey responses, then discuss design choices, and the significance of ASC$^2$End in the finance domain.

\section{The ASC$^2$End System}

The ASC$^2$End system provides insights by comparing a given text corpus against a set of user-defined criteria. This comparison evaluates the relevance of each document in the corpus to a user-defined topic. The ASC$^2$End system is built using abstractive summarization, RAG, binary QA tasks, and reasoning tasks; tasks which have been recently studied to be effective when deployed with generative LLMs \citep{Touvron_2023}. Zero-shot prompting is implemented to enable more flexibility across applications as it removes the need for model finetuning. Our system presents a new approach to the pre-retrieval step of the RAG process through the application of abstractive summarization to relieve token usage while retaining semantic context for extended input contexts.

\begin{figure*}
    \centering 
    \includegraphics[width=1\textwidth]{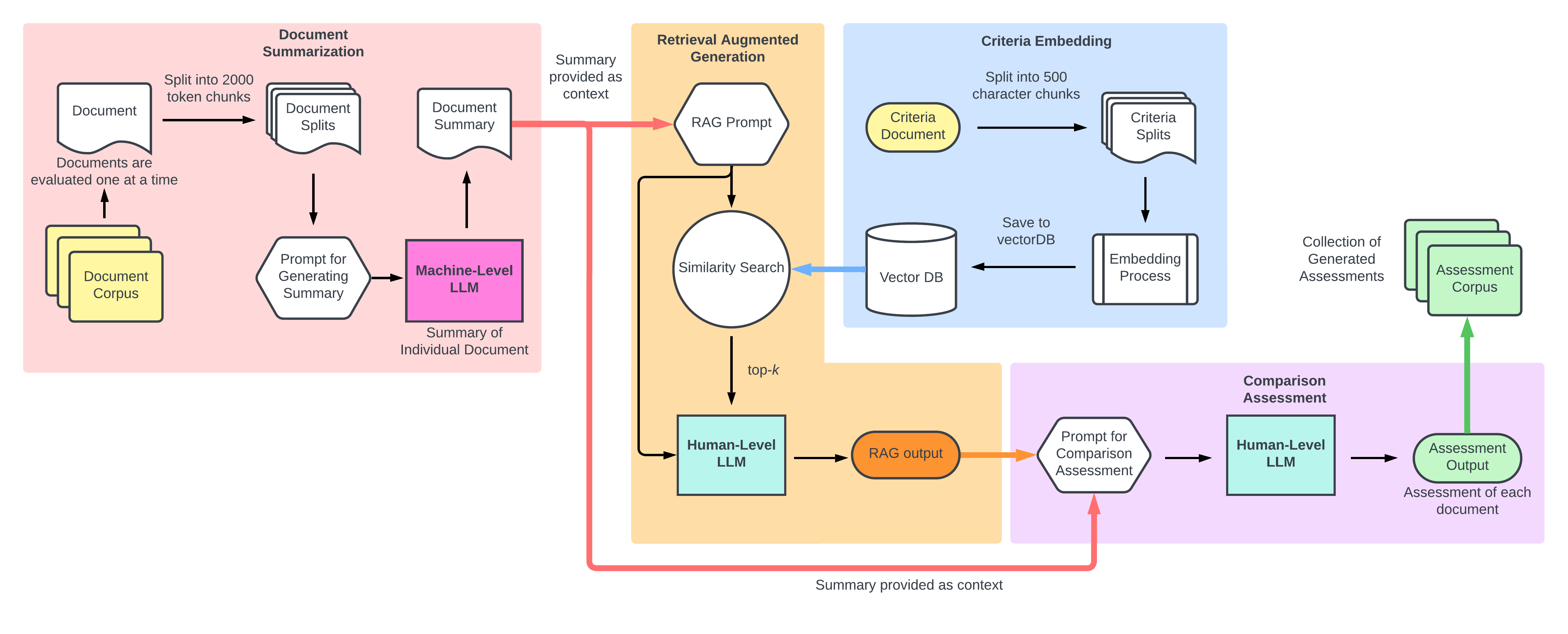}
    \vspace{2cm}
    \caption{ASC$^2$End Pipeline. \textbf{The Document Summarization (DS)} and \textbf{Criteria Embedding (CE)} modules preprocess the information provided by the user. Inputs are highlighted at the beginning of each module (document corpus in DS and criteria document in CE).  The document summary is supplied to both the \textbf{Retrieval Augmented Generation (RAG)} prompt and the \textbf{Comparison Assessment (CA)} module. The vector database is used to drive the similarity search in the RAG module. The RAG prompt is combined with the results of the similarity search, where the information is relayed to a human-level LLM to enhance the retrieved passages. The CA module uses the information preprocessed from the DS and CE modules to perform the comparison assessment task using the same human-level LLM as the RAG process. Highlighted in the final step are the generated assessments for each document.
    } 
    \label{fig_mom0}
\end{figure*}

The ASC$^2$End system comprises four modules (Figure \ref{fig_mom0}) that process the user-specified information and generate the comparison assessment. The first two modules, \textbf{Document Summarization (DS)} and \textbf{Criteria Embedding (CE)}, perform the input processing of the given text corpus and user-defined criteria respectively. The third module, \textbf{Retrieval Augmented Generation (RAG)}, facilitates the similarity search to retrieve and output the relevant information for the next module. The last module, \textbf{Comparison Assessment (CA)}, performs the comparison tasks with the preprocessed data as input. The DS module iteratively performs abstractive summarization on individual documents from the given text corpus to generate a summary for each document. The DS module functions as a pre-retrieval step for the RAG module to distill relevant context efficiently. The CE module vectorizes and splits the user-defined criteria document, storing the segments in a vector database. The RAG module uses the vector database to drive a similarity search between the summary from the DS module, combined with the RAG prompt and the user-defined criteria to return the top-\textit{k} passages. These passages are fed in with the RAG prompt to a human-level LLM to provide an augmented output of these passages to the CA. In the last step, the CA module uses the information provided by the RAG module and the summary from the DS module to generate an output comparison assessment.

\section{Method}

ASC$^2$End's ability to perform large-scale comparisons is facilitated through the system's data preprocessing steps on a given text corpus and set of user-defined criteria. To validate the performance of our system, we sourced a financial news dataset as the candidate text corpus and selected a relevant criteria document as the user-defined criteria.

Model selection, methods of performance evaluation, and experimental setup are described for each data preprocessing (DS and CE), RAG, and CA module. We categorize LLMs by machine or human-level reasoning to assign appropriate models to each ASC$^2$End module. To evaluate the performance of abstractive summarization done by machine-level LLMs, the ROUGE (Recall-Oriented Understudy for Gisting Evaluation) metric \citep{Lin_2004} was used. The performance of the comparison assessment done by human-level LLMs was evaluated using survey participants to optimize the output coherence, quality of the response, and accuracy of the retrieved information.

\subsection{Machine-Level vs. Human-Level Reasoning}

In this study, we divide reasoning tasks into machine-level and human-level based on the level of existing research and the overall task complexity. \citet{Brown_2020} found that larger parameter models are more likely to generate human-like responses, which reflected human-like reasoning when providing solutions to tasks, thus we used parameter sizing as our deciding factor for the two types of reasoning tasks.

Here, machine-level reasoning tasks are defined as those that LLMs perform successfully, as demonstrated by existing studies. The performance of these tasks can be evaluated using NLP precision metrics, such as ROUGE and BLEU scoring without the need for human intervention or validation \citep{Rath_2023}. Examples include NER and abstractive summarization tasks that require simple relation extraction abilities from the LLM.

Human-level reasoning tasks require several different solving strategies when provided to a model or human \citep{Johnson-Laird_2010} and result in a response that reflects chain-of-thought reasoning when arriving at a final answer. Human-level reasoning tasks also require the LLM to understand the information semantically, such that the desired output is acquired. Additionally, the performance of these tasks requires human validation and feedback to ensure final output coherence. Examples of human-level reasoning include performing STS and text comparison tasks and generating responses from criteria-driven prompts.

\subsubsection{\textbf{Model Selection}}

To select the optimal LLMs for the machine and human-level reasoning tasks, the performance of several generative LLMs were evaluated. Answers and scores provided by generative LLMs are prone to hallucinations \citep{Bang_2023}\citep{Huang_2023} and imperfect logic during model inference. However, we addressed this issue by assigning the appropriate generative LLMs for different reasoning tasks. Additionally, the \cite{openleaderboard} leaderboard shows a strong correlation of model size to its performance on test bench tasks, further justifying our definitions of machine-level and human-level tasks. In our study, we compared the performance of the following models: Llama 2 7B, 13B, and 70B \citep{Touvron_2023}, Mistral 7B \citep{A.Jiang_2023}, GPT 3.5 \citep{Brown_2020} and GPT 4 \citep{OpenAI_2023}.

Llama-2 7B, Llama-2 13B, Mistral 7B, and GPT 3.5 were chosen to execute the machine-level reasoning task based on their cost-efficient performance for their parameter size \citep{openleaderboard}. The threshold for determining the reasoning power of the models was chosen based on the trend notice from \cite{openleaderboard} between the training parameter size and expected model performance. Llama-2 13B was specifically chosen as the sole 13B model to evaluate the effects of larger parameters on the summarization task. GPT 3.5 was chosen based on its significant performance difference in reasoning tasks compared to GPT 4 \citep{Espejel_2023}, as machine-level reasoning tasks do not require as much reasoning power. This study was performed before the existence of GPT 4o, thus, its performance was not evaluated.

Human-level reasoning models chosen for the experiment were Llama-2 70B and GPT 4. These are currently, at the time of writing, the most competitive LLMs to be used for complex-level reasoning while fitting within our experimental constraints. Larger parameter models can generate human-like text \citep{Brown_2020}, reflecting that these models are better suited to elicit human-like reasoning. Llama-2 70B is currently the best-performing 70 billion parameter open-source base model available \citep{openleaderboard} and GPT-4 has studies demonstrating performance comparable to human-like reasoning \citep{Bubeck_2023}\citep{OpenAI_2023}\citep{H.Liu_2023}.

To evaluate the performance of the machine-level LLMs in the DS module, we utilized the ROUGE metric. ROUGE scoring is widely used in many summarization-focused studies involving generative LLMs \citep{S.Liu_2023}\citep{Kumar_2023} as it provides a simple method of measuring the amount of information retained from the source text. The ROUGE score determines the similarity between a summary and a reference text based on text overlap. Scoring generative LLMs with ROUGE does not reflect the readability of the model's response. Therefore, the aim of using ROUGE was to assess each model's ability to retain information from the source document systematically.

 \begin{table}[h]
\centering
\caption{Questions presented to the survey participants to evaluate the quality of the comparison response. Survey participants were asked to assign a value of 0 (no) or 1 (yes) to each question indicating if they agreed that the comparison response satisfied the conditions. 
}
\begin{tabular}{l} 
  \hline
 Questions\\ 
  \hline
 Roles of Participants Stated \\ 
 Transaction Type Identification \\ 
 Transaction Amount (\$) Identification \\
 Comparison to sustainability criteria is justified \\
 Do you agree with the confidence score \\
 \& explanation? \\
  \hline
\end{tabular}

\label{Table1}
\end{table}

Evaluation of the human-level LLMs' performance in the comparison assessment was a complex task that required the validation of output readability and correct process reasoning. Therefore, we collected human feedback through the creation of a survey. A dataset was created by randomly selecting and masking 20 model output pairs that correspond to the same document summary, with one of each pair generated by either the GPT-4 or Llama-2-70B model to allow for comparison of performance between the models. As a result, the survey comprised of ten GPT-4 response outputs and ten Llama-2-70B response outputs. The same survey was distributed to all participants for consistency. Randomly sampling the model output dataset may introduce bias in the results. However, since the models were being evaluated comparatively, the empirical scores had a lower impact on the overall analysis. 

The survey used a 5-point scoring system by asking participants to answer 5 questions per document entry, as shown in Table\ref{Table1}. Each question was scored either a 0 (no) or 1 (yes). The first three tasks evaluated the ability to retrieve information from the input context and the last two tasks evaluated the reasoning abilities of the selected LLM. This scoring method was chosen to lower answer scoring ambiguity. Each participant was provided with an explanation of their task along with the user-defined criteria document. Participant data was completely masked for confidentiality and participants had no formal domain knowledge in finance or sustainability.

\subsection{\textbf{Data Preprocessing of Text Corpus and Criteria}}

The ASC$^2$End system performs preprocessing on both the input text corpus and the set of user-defined criteria. The text corpus used in this study was financial data sourced from a news database to demonstrate the feasibility of using raw, unstructured data. 
Document summarization was performed with generative LLMs to process the raw financial data. The user-defined criteria document was uploaded to a vector database so that a similarity search could be conducted. 

\subsubsection{Data Acquisition}

The data search scope for this study was to analyze a financial institution's publicly available transaction records (i.e., the text corpus) and compare them with the institution's sustainable finance guidelines (i.e., user-defined criteria). The eligible years of analysis were from 2020-2022 given the availability of sustainable finance reports from the financial institution. Any of those three years could have been used as the ASC$^2$End system was designed to enable frequent data analyses without limitations. We chose 2021 for its sufficiently large dataset to test our system's capabilities.

Data for the calendar year 2021 was obtained from Factiva, a financial database owned by Dow Jones \& Company \citep{factiva}. Factiva compiles articles from various open-source news outlets and information posted on the SEC EDGAR database. Web scraping was employed to extract the article title and document contents. The resultant text corpus was a 2-column CSV file with 1253 entries, the same number of available articles from 2021 on Factiva. These articles had an average of 7500 words. The user-defined criteria was a 20-page PDF document extracted from a financial institution's website outlining sustainable finance guidelines.

\begin{figure}[h]
     \centering 
    \fbox{\begin{minipage}{22em}
    Query:
    Given this text: \textbf{\{split\_text\}}...
    generate a TL;DR.
    
    Guidelines for your answer:
    
    1. Include all detailed information relevant from the text.
    
    2. Formulate concise answers, grounded on facts from context. Keep answers logical.
    
    3. Use point form answers.
    
    Answer: TL;DR:
    \end{minipage}}
    \vspace{0.6cm}
	\caption{Prompt used to perform the summary task. \{split\_text\} refers to each 2000 token chunk that is provided to the LLM to summarize.} 
	\label{fig_mom1}%
\end{figure}

\subsubsection{Document Summarization (DS)}

The Document Summarization (DS) module employed Abstractive Summarization to process the financial text corpus and generate standardized-length summaries. Abstractive summarization with the use of generative LLMs has been well-documented in several studies \citep{S.Liu_2023}\citep{Kumar_2023}. \citet{S.Liu_2023} performed abstractive summarization using lower complexity models. Thus, we determined that the use of machine-level generative LLMs would be suitable for this task.

Tokens were measured at 4 English characters per token \citep{tokenizer}. In our study, we preprocessed the obtained data by splitting each document into 2000 token chunks, with each chunk summarized into 250 token segments. The summarized segments were concatenated to form the final document summary. Since the smallest context window of the models used in experimentation was 4096 tokens, we found that 50\% of the total context or 2000 tokens, was the maximum length of each chunk that could be provided without raising runtime errors. The 250-token length for summarized segments was chosen based on previous work done by \citet{Shapira_2018}, which showed that gains to ROUGE scoring had diminished returns past the 200-word range. 250 tokens is approximately 190 words and 12.5\% of the original 2000-token chunk length. Figure \ref{fig_mom1} illustrates the prompt used to generate the summary for each 2000-token chunk. \citet{Radford_2019} found that using the "TL; DR" token in the prompt was an effective method to incite a summary response. Additionally, we specified guidelines in the prompt to control the generated response structure (e.g. "Formulate concise answers") such that tokens were used effectively to retain information.

The maximum threshold length of the generated summaries was determined to be 1250 tokens to account for the minimum context window length of the LLM performing the comparison assessment as well as the hardware limitations of the experimental setup. If the final document summary was longer than 1250 tokens, or five 250-token segments, then the DS process was repeated until the output summary length was below the threshold. Previous experiments were conducted with a higher threshold length of 2500 tokens per generated summary to take full advantage of the context window length. However, the comparison task could not be performed due to memory limitations, thus the threshold size of the document summary was reduced to 1250 tokens, or 30\% of the total context (i.e. 4096 tokens). Results from both the original 2500 token threshold and the new 1250 token threshold were explored to discuss the effects of shortening the available context space for abstraction summarization. We hypothesized that the decreased context window would negatively affect the performance of the DS module.
% All aspects of each supplied document are important as they maintain the semantic meaning and context of the documents. However, when using criteria for comparison, only relevant sections of the criteria need to be recalled to conduct a successful comparison assessment.

\subsubsection{Criteria Embedding (CE)}
The user-defined criteria document was embedded and stored in a vector database such that the user could retrieve passages that were most relevant to the input document summary. The criteria document was vectorized and split into 500-character (125 tokens) chunks with a 20-character overlap. It was important to have character overlap when splitting the criteria for coherent retrieval. The criteria document split size was chosen based on the experimentation described in chatReport \citep{Ni_2023}, which stated that a split size of 500 characters and 20-character overlap resulted in the best retrieval performance. 

\begin{figure}[h]
     \centering 
    \fbox{\begin{minipage}{22em}
    Query:
    
    Given this document delimited by "": "\textbf{\{summary\}}":
    Provide the most relevant information only from the criteria that matches with the given document in terms of \textbf{\{target\_topic\}}?

    Answer:
    \end{minipage}}
    \vspace{0.6cm}
	\caption{Prompt provided to perform the RAG task. The information provided to \{summary\} was each summarized document and \{target\_topic\} was a user-defined input to control the scope of the search.} 
	\label{fig_mom2}%
\end{figure}

\subsection{\textbf{Retrieval Augmented Generation (RAG)}}
% By conducting a similarity search between the vector database and an individual document of the corpus
%and returning the most similar passages compared to the input document summary. 
RAG retrieves relevant passages from the vector database by conducting a semantic search. The prompt given to the RAG task (Figure \ref{fig_mom2}) combines the input document summary with a simple QA query. The document summary was integrated into the RAG prompt to provide the necessary context and function as a pre-retrieval query transformation. Human-level LLMs were chosen to perform the RAG task as we required the LLM to perform STS with the retrieved passages and the supplied document summary to return the most relevant information.

The user can designate the focus of the search by defining a target topic that directs content retrieval. A semantic search was performed by applying the RAG prompt to the vector database to find the top-\textit{k} = 3 passages. \textit{k} is a tunable hyperparameter that is dependent on the specified application and set of criteria. In this study, it was found that beyond \textit{k} = 3, the retrieval response did not capture additional information. The top-\textit{k} passages retrieved by the semantic search are an intermediary output that is presented alongside the RAG prompt as additional context for a human-level LLM to answer the query. The output response of the human-level LLM is an initial comparison that is informed by relevant retrieved passages from the criteria.

\begin{figure*}[h]
    \centering
    \resizebox{\textwidth}{!}{\noindent\fbox{\begin{minipage}{40em}
    % \parbox{\textwidth}{%
    Prompt:
    You are an AI model assisting a Financial Analyst at \textbf{\{company\}}. Your task is to analyze the document delimited by "": "\textbf{\{summary\}}" and provide a thorough, yet concise analysis in the following format:

    1. Article Date: [Please input the date of the article here in MM/DD/YYYY format]

    2. Participants of the transaction: [Please provide a brief description of \textbf{\{company\}}'s role in relation to the article, then list the entities involved in the transaction mentioned in the article]

    3. Transaction and Transaction type: [Please indicate whether a transaction has taken place. If yes, state the type of transaction.]

    4. Transaction amount in dollars: [If a transaction has occurred, please specify the amount in dollars. If no transaction, please input \$0]

    5. Comparison: [Based on the following criteria, delimited by "": "\textbf{\{retrieved\_text\}}". Provide a concise comparison between the document and provided criteria and 
    discuss the relevancy of the document to \textbf{\{target\_topic\}}. Use specific information from the criteria and be very critical in your assessment].

    6. Confidence score: [Please provide a score between 0-100 indicating the degree to which the document discusses topics related to \textbf{\{target\_topic\}}. A score of 0 means the document is not at all related to \textbf{\{target\_topic\}}, a score of 50 means there are many uncertainties as to its correlation to \textbf{\{target\_topic\}}, and a score of 100 means the document content is entirely about \textbf{\{target\_topic\}}. If the transaction amount is \$0 or there is no transaction, please input a score of 0. Use your comparison to affect your decision, skepticism and implicit assumptions in the answer needed to negatively affect the confidence score.]
    
    Please remember to:
    
    1. Provide factual and concise answers. 2. Critically evaluate the information from the document. 3. Use bullet points for your answers. 4. Do not explain your thought process. 5. Do not include extra text in addition to your analysis outside of the six points of analysis. 6. "document" should only refer to the provided article document. 
    
    Response:
    \end{minipage}}}
    \vspace{1.5cm}
    \caption{Prompt provided to perform the comparison assessment. The \{summary\} generated from the Document Summarization module and the \{retrieved\_text\} outputted from the RAG module are provided to the prompt to perform the information retrieval and comparison tasks. The \{target\_topic\} and \{company\} are provided by the user to direct the scope of comparison.}
    \label{fig_mom3}%
\end{figure*}

\subsection{\textbf{Comparison Assessment (CA)}}

The Comparison Assessment (CA) module performs entity recognition, information extraction and comparison assessment tasks through the use of generative LLMs. In this experiment, the information extraction tasks included the identification of financial transaction details such as whether a transaction occurred, the transaction amount and type, and the relevant participants. The main task of the CA module is the comparison between the RAG output (i.e. relevant passages of the criteria) and the generated document summary. Complimentary to the comparison, a confidence score was assigned to the relevancy of the document summary against a user-defined topic. The tasks defined in the CA module are a mix of machine-level and human-level reasoning tasks, with the most important task being the comparison of information itself. To ensure the success of the comparison task, we determined that the CA module was performed using human-level LLMs.

To drive the outlined tasks in the CA module, the comparison prompt in Figure \ref{fig_mom3} was provided to an LLM. The comparison prompt implements multiple zero-shot prompting strategies to extract relevant information from the input context and to conduct the comparison tasks. These strategies include explicit examples, guideline specifications for finding the relevant information, and rules to direct the model response. We specifically used the phrase "Do not explain your thought process" as one of the rules to remove words that had no relevance to the comparison. Additionally, we wanted to investigate if the language model could successfully perform the task without explicitly stating its chain of thought.

%The generated output from RAG was combined with the respective summary such that a generative LLM performed the comparison assessment.
%A scoring rubric (Table \ref{Table1}) was developed to assess the model's ability to complete the assessment tasks within the comparison.

\begin{figure*}[h]
     \centering 
	\includegraphics[width=1\textwidth]{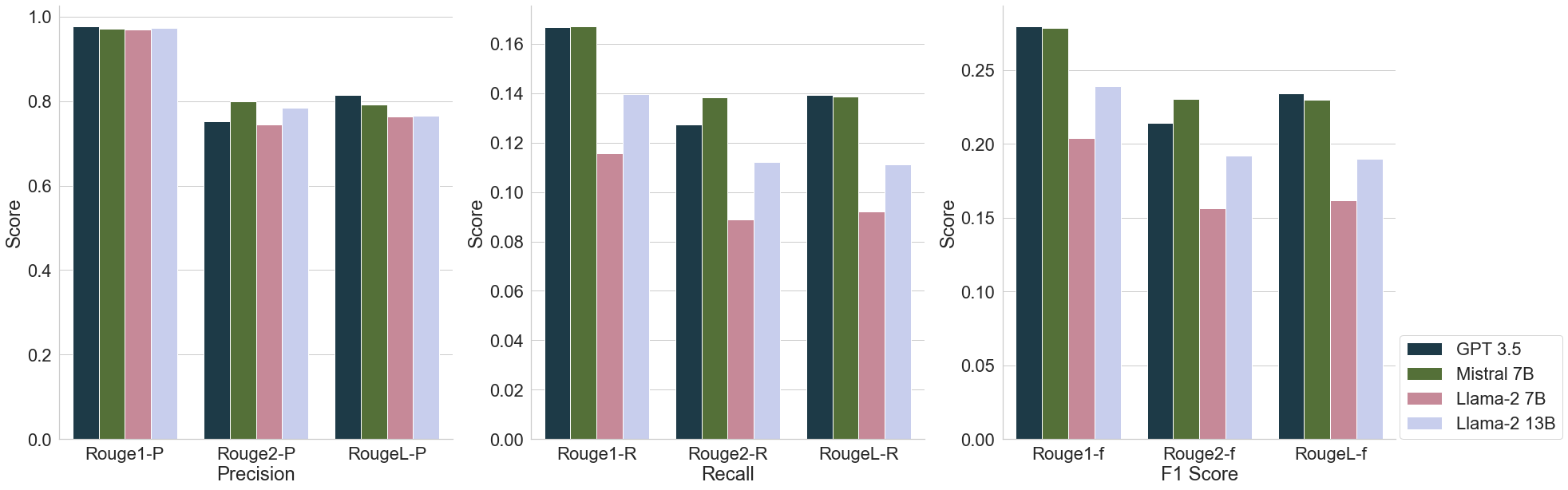}
    \vspace{0.5cm}
	\caption{ROUGE Scores of the first abstractive summarization assessment with a 2500 token limit of the four machine-level reasoning LLMs. n = 1,2, L were the evaluated n-grams for the ROUGE scoring, scores range from 0-1. The F1 score (right) was calculated with the influence of the precision (left) and recall (middle) scores.
 } 
	\label{fig_mom4}%
\end{figure*}

\subsection{Baseline and Ablation Study Experiment Setup}

To demonstrate the power and advancement of the ASC$^2$End system, we designed a baseline comparison and ablation study to compare the accuracy and clarity of the results to our system. The ablation study highlighted the importance of each module of the ASC$^2$End system. Additionally, we compared the token cost difference between the various methods. Recent developments in LLMs have yielded models with greatly increased input context(128k tokens) enabling these experiments. The baseline experiment was supplied with the unprocessed news document, the whole criteria document, and identical prompts used in the ASC$^2$End system to generate responses. The ablation study was performed with the section of focus removed from the ASC$^2$End pipeline. GPT 4o was applied to the original model, baseline and all ablation studies to maintain testing consistency and comparison. Only 50 randomized data points, a small subset of the original dataset, were used for the baseline and ablation study.

\subsection{\textbf{Experimental Setup}} 

The experiment was completed on an A6000 GPU with 48 GB of RAM. Open-source models were loaded into local memory using the GPTQ method \citep{Elias_2022} where the model parameters were quantized to decrease the total memory load. Model hyper-parameters are shown in Table \ref{Table2}. Models were set to a temperature of 0 to maintain the results' reproducibility and keep output formats consistent for all candidate documents. ASC$^2$End was developed using the LangChain \citep{langchain} framework for all API calls and for managing the vector database requests. The embedding model used for the vector database was the Beijing Academy of Artificial Intelligence (BAAI)’s “bge-base-en-v1.5” \citep{Xiao_2023} as it was the best performing open-source embedding model for the model size. Additionally, the BAAI model outperformed OpenAI’s paid “text-embedding-ada-002” model on the Massive Text Embedding Benchmark (MTEB) leaderboard \citep{Muennighoff_2022}. The RAG task for this system is powered by FAISS (Facebook AI Similarity Search) \citep{Johnson_2017} acting as the vector database, due to its efficiency in performing similarity search.

\begin{table}[h]
\centering
\caption{Hyper-parameters used for each model. Models were divided into Machine-level LLMs (GPT 3.5, Mistral 7B, Llama 2 7B, 13B) and Human-level LLMs (GPT 4, Llama 2 70B).
}
\begin{tabular}{l c c} 
  \hline
 LLM Class& Temperature & Max New Tokens \\ 
  \hline
 Machine-level &  0 & 250 \\ 
     Human Level & 0 & 500 \\ 
  \hline
\end{tabular}
\label{Table2}
\end{table}

\section{Results}

In this section, we discuss the model performances on both the abstractive summarization task and the comparison assessment task evaluated using the ROUGE score and survey answers, respectively. Then we present the performance of the ASC$^2$End framework to baseline methods and perform an ablation study to emphasize the impact of each defined module.

\begin{figure*}[h]
     \centering 
	\includegraphics[width=1\textwidth]{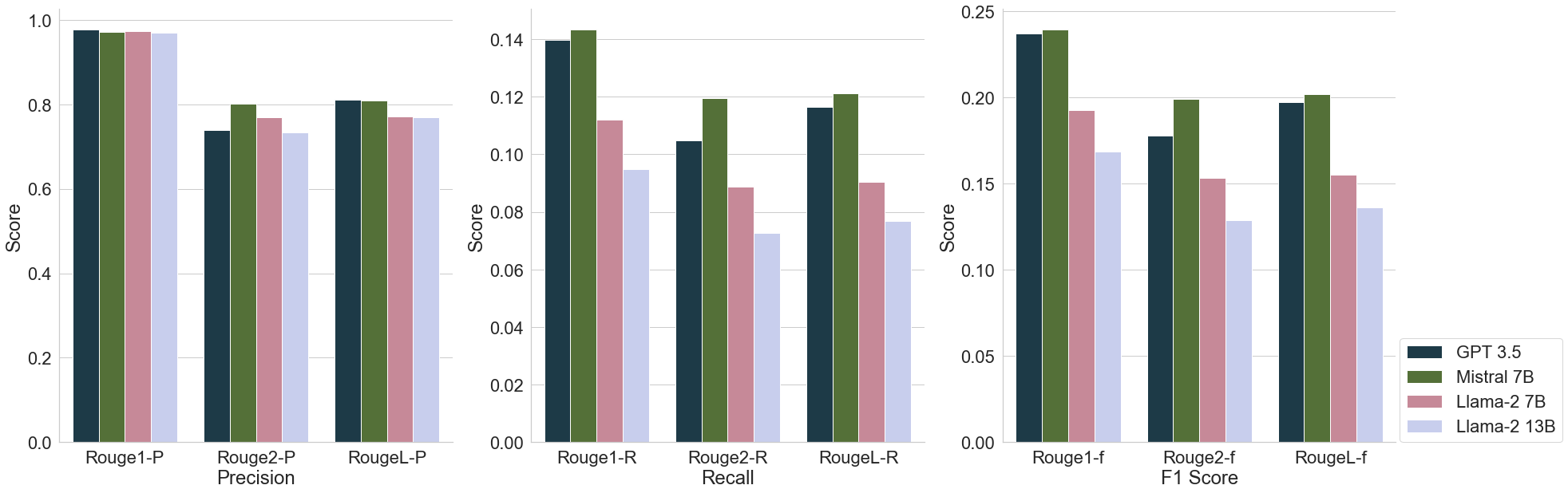}
    \vspace{0.5cm}
	\caption{ROUGE Scores of the second summarization assessment with a 1250 token limit of the four machine-level reasoning LLMs. n = 1,2, L were the evaluated n-grams for the ROUGE scoring, scores range from 0-1. The F1 score (right) was calculated with the influence of the precision (left) and recall (middle) scores.
 } 
	\label{fig_mom5}%
\end{figure*}

\subsection{\textbf{Abstractive Summarization Results}}

The scores presented for abstractive summarization are the averaged ROUGE values across the given corpus of 1253 documents. There are two sets of summarization results, the first using a maximum output length of 2500 tokens and the second using 1250 tokens. Both sets of results are presented to highlight different relative model performances on a limited context window.
% Overall, shrinking the maximum output length negatively impacted the quality of information generated.

ROUGE \citep{Lin_2004} is calculated based on the text overlap using the calculated precision and recall values. The score ranges from 0 to 1, where 0 indicates poor similarity and 1 indicates strong similarity between the summary and reference text. ROUGE can be evaluated with any number of n-grams. The chosen evaluations for this experiment were unigram (n=1), digram (n=2), and the longest sub-sequence of text (n=L) for evaluation. Unigram performance measures the number of single words that match the original document, reflecting how overall information was retained. Conversely, digram and sub-sequence performance measure the quality of semantic meaning retained in the summary. The ROUGE formulas for calculating the unigram \& digram and the L-scores are presented in Figures \ref{fig_mom6} and \ref{fig_mom7} respectively.

In the 2500-token length summarization results (Figure \ref{fig_mom4}), unigram (n=1) precision across all four machine-level models was competitive with scores close to 1. A significant decline in performance in both precision and recall was observed when the remaining n-gram performances were compared to the unigram (n=1) performance. Due to paraphrasing, word-pair overlaps and sub-sequence matching were not maintained in abstractive summarization, leading to lower scores in digram (n=2) and subsequence (n=L) scores. The recall score is based on the amount of text overlap between the candidate and reference text compared to the total length of the reference text. Summarization shortened the length of the source documents to less than 12.5\% of its initial length, and a shorter context length meant less text to compare for the recall score. Thus, the recall score in every case for all four models was significantly lower than the precision scores. In terms of models, Mistral 7B displayed a strong score in the ROUGE-2 scenario, beating GPT 3.5 and the Llama models (Figure \ref{fig_mom4}). Mistral 7B also has competitive results compared to GPT 3.5 in ROUGE-1 and ROUGE-L scoring and is an open-source model, making it an economical option for larger-scale workflows. It was also observed that there were significant performance decreases in the recall scores of the Llama models. Llama model outputs are more lengthy and are more likely to generate sequences of text that may not be relevant to the summary, thus negatively affecting the recall score.

\begin{figure}
    % {\setlength{\mathindent}{0cm}
    \begin{equation}
        \textnormal{ROUGE-1}_\textnormal{precision} = \frac{\textnormal{cand. unigram} \cap \textnormal{ref. unigram}}{|\textnormal{cand. unigram}|}
        \label{eqn1}
    \end{equation}
    % }
    % {\setlength{\mathindent}{0cm}
    \begin{equation}
        \textnormal{ROUGE-1}_\textnormal{recall} = \frac{\textnormal{cand. unigram} \cap \textnormal{ref. unigram}}{|\textnormal{ref. unigram}|}
        \label{eqn2}
    \end{equation}
    % }
    \begin{equation}
        \textnormal{ROUGE-1}_\textnormal{F1} = 2\cdot\frac{\textnormal{precision}\cdot \textnormal{recall}}{\textnormal{precision} + \textnormal{recall}}
        \label{eqn3}
    \end{equation}
    \vspace{0.5cm}
    \caption{ROUGE-1 Equations. Candidate text (cand.) refers to the summarized document, and the reference text (ref.) refers to the original text. ROUGE-2 uses digrams instead of unigrams}
    \label{fig_mom6}%
\end{figure}

The 1250-token length summarization task (Figure \ref{fig_mom5}) was conducted  {to ensure runtime success} and to analyze the losses in ROUGE scoring for a shorter output summary length. A direct comparison in the ROUGE values between the two summarization tasks was made in Table \ref{Table3}. The precision of all four models was almost identical to the first summarization task, implying that the context of the information did not change during the shortening of the summaries. The recall scores decreased by at least 2\%, with the greatest decrease in recall for the Llama-2 13B model. As a result, the larger parameter Llama model was deemed not suited for summarization tasks with token limitations due to its verbosity. Llama-2 7B had similar scores across both summarization tasks but still performed significantly worse than the top models. From this 1250-token length experimentation, we found that the top-performing models from the first summarization task experienced negligible changes in performance despite the shortened summary length. Mistral 7B outperformed GPT 3.5 in every recall measure and had improved overall F1 scoring. In addition, Mistral 7B recall scores did not take a severe performance hit in the second summarization task. Therefore, Mistral 7B was the most suitable machine-level LLM to employ in the DS module for running abstractive summarization.

The results of the second summarization task were promising because of the minor shifts in ROUGE precision scores implying that semantic context was still maintained. The shorter context demonstrates that using abstractive summarization with different context lengths may not negatively affect the overall performance of the pre-retrieval step in the ASC$^2$End system. The shorter context also supports our aim to address scaling the process for larger datasets, as shorter output contexts are more financially and computationally efficient. The investigation of several prominent LLMs in applying the ASC$^2$End framework demonstrates a preliminary benchmark that is beneficial in understanding the expected performance of similarly-sized models and helps with model selection for future implementation.

\begin{table*}[h]
\centering
\caption{ROUGE scores of the first and second summarization experiments. The model performance from the first and second summarization assessments are indicated by -1 and -2, respectively. Bolded values reflect the best score in the second summarization assessment.
}
\resizebox{\textwidth}{!}{\begin{tabular}{l c c c c c c c c} 
  \hline
 & GPT 3.5-1 & GPT 3.5-2 & Mistral 7B-1 & Mistral 7B-2 & \begin{tabular}{@{}c@{}}Llama-2 \\ 7B-1\end{tabular} & \begin{tabular}{@{}c@{}}Llama-2 \\ 7B-2\end{tabular} & \begin{tabular}{@{}c@{}}Llama-2 \\ 13B-1\end{tabular} & \begin{tabular}{@{}c@{}}Llama-2 \\ 13B-2\end{tabular} \\ 
  \hline
Precision &&&&&&&&\\
n = 1 & 0.978 & \textbf{0.978} & 0.971 & 0.972 & 0.970 & 0.974 & 0.973 & 0.970 \\
n = 2 & 0.751 & 0.740 & 0.799 & \textbf{0.801} & 0.744 & 0.770 & 0.783 & 0.734 \\
n = L & 0.815 & \textbf{0.812} & 0.793 & 0.808 & 0.763 & 0.772 & 0.766 & 0.770 \\
  \hline
Recall &&&&&&&&\\
n = 1 & 0.167 & 0.140 & 0.167 & \textbf{0.143} & 0.116 & 0.112 & 0.139 & 0.095 \\
n = 2 & 0.127 & 0.105 & 0.138 & \textbf{0.120} & 0.089 & 0.089 & 0.112 & 0.073 \\
n = L & 0.139 & 0.116 & 0.139 & \textbf{0.121} & 0.092 & 0.090 & 0.111 & 0.077 \\
  \hline
F1 &&&&&&&&\\
n = 1 & 0.280 & 0.237 & 0.278 & \textbf{0.240} & 0.204 & 0.193 & 0.239 & 0.169 \\
n = 2 & 0.214 & 0.178 & 0.230 & \textbf{0.199} & 0.157 & 0.153 & 0.192 & 0.129 \\
n = L & 0.234 & 0.197 & 0.230 & \textbf{0.202} & 0.162 & 0.155 & 0.190 & 0.136 \\
  \hline

\end{tabular}}
\label{Table3}
\end{table*}

\begin{figure}
    \begin{equation}
        \textnormal{ROUGE-L}_\textnormal{precision} = \frac{LCS(\textnormal{cand., ref.})}{\textnormal{\#words in cand.}}
        \label{eqn4}
    \end{equation}
    \begin{equation}
        \textnormal{ROUGE-L}_\textnormal{recall} = \frac{LCS(\textnormal{cand., ref.})}{\textnormal{\#words in ref.}}
        \label{eqn5}
    \end{equation}
    \begin{equation}
        \textnormal{ROUGE-L}_\textnormal{F1} = 2\cdot\frac{\textnormal{precision}\cdot \textnormal{recall}}{\textnormal{precision} + \textnormal{recall}}
        \label{eqn6}
    \end{equation}
    \vspace{0.4cm}
    \caption{ROUGE-L Equations. Candidate text (cand.) refers to the summarized document, and the reference text (ref.) refers to the original text.}
    \label{fig_mom7}%
\end{figure}

\begin{table*}[h]
\centering
\caption{Obtained results of survey participants for the comparison assessment. Survey participants assigned each question a value of 0 or 1  for no or yes, respectively, based on whether they agreed that the response satisfied the questions. Scores presented are the average of results from 21 survey participants.
}
\resizebox{\textwidth}{!}{\begin{tabular}{l c c c c c c} 
  \hline
& Correctly Stated & Correctly Stated & Correctly Stated & Correct & Correct Confidence & Overall \\
    & Roles & Transaction Type & Transaction Amount & Comparison & Score & Score \\
  \hline
 Llama\,2 &  0.760 & 0.875 & 0.825 & 0.432 & 0.562 & 3.453 \\ 
 GPT-4 & \textbf{0.893} & \textbf{0.925} & \textbf{0.830} & \textbf{0.698} & \textbf{0.810} & \textbf{4.155} \\ 
  \hline
\end{tabular}}

\label{Table4}
\end{table*}

\subsection{\textbf{Comparison Assessment (CA) Results}}

When evaluating the human-level LLMs performing the CA tasks, we prioritized the clarity of model output, accuracy of model response, and the presentation of information. The scores presented in the comparison assessment reflect the human-annotated scores of 21 survey participants. For this section, Llama 2 70B will be called Llama 2.

The averaged survey scores for each model are shown in Table \ref{Table4}. GPT 4 performed better than Llama 2 by 0.7 points on average and scored higher in every category. GPT 4 scored 0.2 – 0.3 points greater than Llama 2 in assessing the comparison and providing reasoning to justify its confidence score. Score disparity may be attributed to the different reasoning abilities of the two models. Scores on identifying stated information were more similar between Llama 2 and GPT 4, with a 0.05 - 0.1 point difference between the first three categories focusing on retrieving information from the input context. The reason for the similar scores is attributed to the lower task complexity of NER tasks.

According to the survey results, there are major differences between the human-level reasoning abilities of the models. Llama 2 obtained an unimpressive score of 0.432 when evaluating the correct comparison given a candidate summary and the retrieved passages from the user-defined criteria. Llama 2 struggled to identify the sentiment of the candidate text and it conformed the candidate sentiment to the sentiment of the retrieved reference text. The model's inability to perform the comparison task resulted in misidentifying and hallucinating the candidate text's conformity to the target topic. Similarly, when providing reasoning for its confidence score on the target topic, Llama 2 obtained a low score of 0.56.

On the other hand, GPT 4 excelled in justifying its generated confidence score, with an average score of 0.81. This score indicates that the GPT 4 model accurately determined the sentiment of the candidate text and used reasoning to provide a valid response. The score also reflected that GPT 4 explained its reasoning succinctly, so the survey participants agreed with the response. GPT 4 obtained a score of 0.698 for its ability to identify the correct comparison. Compared to Llama 2’s 0.431, GPT 4 was better at accurately identifying sentiment and making accurate comparisons between the candidate and reference texts. 

The comparison score reflects the practical capabilities and quality of responses generated by the ASC$^2$End system. The performance disparity between GPT 4 and Llama 2, as seen in Table \ref{Table4} demonstrates the importance of model selection and the complexity of the comparison task. The GPT 4 results show that our system successfully implemented a text comparison solution, and the Llama 2 results show clear shortcomings in a cohesive text comparison. The text comparison task is a complex operation that requires the LLM to "understand" the provided texts and generate relevant comparisons and discussions. The GPT 4 result indicates that the pre-retrieval RAG process with abstractive summarization provided relevant information to the model and enabled full function of the ASC$^2$End system.

\subsection{Baseline Comparison}

The baseline experiment of performing the comparison task was completed using GPT 4o. The baseline experiment was impossible to perform until the existence of larger context models. We scored the performance of the baseline model based on its accuracy to provide specific, context-driven comparisons in its response. The results generated from the baseline showed expected success in completing NER tasks such as date and company recognition driven through prompt engineering. The test data sample included one positive document that fit the targeted search focus, which the baseline successfully identified. However, it was observed that the dates extracted in the baseline experiment were associated with the published date, not the transaction date mentioned within the article. For runtime and token usage performance, the baseline model required 825\% more tokens than the ASC$^2$End model with only a decrease in runtime by 3.8\%, summarized in Table \ref{Comp_table}. The baseline experiment had an accuracy of 80\%, 14\% lower than the ASC$^2$End model in evaluating the comparison tasks. The comparison generated by our system was more concise and specific compared to the baseline experiment. Specific comparison topics regarding the context of the article were discussed in the outputs of the ASC$^2$End system but not in the baseline.

The baseline experiment highlighted some future improvements to the ASC$^2$End system, with a focus on stronger specificity in abstraction. Arguably, using abstractive summarization requires another LLM, but it was completed locally in this experiment on an open-source model for a lower financial cost. If desired, summarization could still be done with GPT 4o with fewer tokens than the baseline. Additionally, the difference in token usage is significantly noticeable in larger workflows, creating a much larger financial burden and unnecessarily using computational resources to obtain less desirable results when applying only the baseline experiment. Additionally, SOTA LLMs like GPT 4o may not be feasible for access to perform this comparison task, namely in organizations and academic settings where models have a 4000-token limit. The ASC$^2$End system provides a cost and computationally-effective solution to text comparisons at scale while outperforming the presented baseline.

\subsection{Ablation Study}

Each module of the ASC$^2$End system serves a vital function to generate the final output analysis for each document. To extenuate the importance of each module, we perform several ablation studies on a small subset of data to demonstrate the differences in output quality and token usage in each ablation study. Additionally, we discussed the accuracy of comparison of each ablation experiment with the ASC$^2$End system, outlining the limitations that the ASC$^2$End system addressed in its implementation. These studies were evaluated on the same dataset, model and rationale as the baseline experiment.

\subsubsection{Removal of the DS Module}

By removing the DS module from the system, the performance difference of applying abstractive summarization was investigated. This ablation study had a 25\% increase in runtime and a 291\% increase in token usage compared to our model, seen in Table \ref{Comp_table}. The cause of the increased run time and token usage was due to the varying article lengths, resulting in more tokens used during inference. During accuracy scoring, it was observed that even though 94\% of the articles were correctly identified and analyzed, there was no change in the retrieved information across different articles of focus. The issue was particularly evident with longer context documents, where retrieval struggled to provide robust comparisons to the criteria document. This ablation experiment failed to generate unique comparison points across the tested data points. In our system, the distilled data provided by the DS module effectively directs the RAG search by providing significantly less input context to compare to for retrieval. The effectiveness of providing distilled and directed data is further amplified by the observations made on the output quality of this ablation experiment.

\subsubsection{Removal of the RAG Module}

Removal of the RAG module demonstrated the effects of specific search compared to providing the entire document for comparison. This ablation task used 526\% more tokens than the ASC$^2$End model but had a runtime difference of -21\%, as seen in Table \ref{Comp_table}. 
This study observed that the comparison discussion had a lower accuracy of 78\% compared to our model. This observation can be attributed to using the LLM to search for relevant points of comparison instead of having RAG direct the context for comparison. Removing the RAG module required loading the entire 20-page criteria comparison document for each inference. The accuracy results reflect a gap in non-RAG approaches to direct evidence-based text comparisons and instead output generic results due to extended context. The application of the RAG module in our system provided directed context for the CA module, instead of depending on the LLM itself to locate the relevant information in the provided document for the comparison.

\subsubsection{Removal of the CA Module}

Presently, RAG-focused architecture focuses on different forms of information retrieval to optimally answer the user query in closed domain QA \citep{Pan_2022}\citep{Ram_2023}. Applying RAG is very powerful for obtaining in-context information to enhance the LLM's response, but RAG processes are optimized for only QA-related tasks. The removal of the CA module showed the implications of its effectiveness in our system. For this experiment, we modified the prompt from the CA module and combined it with the RAG prompt to have the RAG complete the CA module's task. From experimentation, this study took 65\% less tokens to complete and 57\% less time when compared to our model, seen in Table \ref{Comp_table}. However, the outputted comparison discussion and the generated confidence score were incorrect and reflected in its 8\% accuracy score. It was observed that the comparison discussion conclusion consistently contradicted the generated confidence score, reflecting a high degree of hallucination not observed in any other experimentation. The hallucination is driven by the retrieved information, not the input context, indicating that a RAG-only pipeline is inadequate for text comparison. In the ASC$^2$End system, The CA module restates the article summary and only queries the RAG module to search for the relevant passages, not conduct the comparison. Within the RAG module, conducting the search and comparison analysis in one step poses the risk of hallucination due to the searched information affecting the LLM's available context.

\begin{table}[h!]
\centering
\caption{Results of ASC$^2$End ablation study on 50 randomly sampled data points. The percentage difference in token usage and runtime compared to our model was used to highlight the differences in model performance. The accuracy of each study was measured by the model's ability to specifically determine the areas of comparison to the criteria document and correctly assign a confidence score based on its comparison. the presented accuracy is the percentage of the 50 test samples properly identified, labelled either 0 or 1 based on specificity and correctness. Raw values for the model are reported to provide the scale of the experiment. *Accuracy defined in the No-DS ablation study does not reflect model reasoning due to identical comparison points for different articles.}
\begin{tabular}{lccc}
 \hline
Description & \% Token Difference & \% Runtime Difference & Accuracy \\
 \hline
Baseline & +824.9\% & -3.8\% & 80\% \\
No DS & +290.8\% & +25.2\% & 94\%* \\
No RAG & +526.4\% & -20.5\% & 78\% \\
No CA & -65.5\% & -57.3\% & 8\% \\
Model & 2277.28 & 6m 5s & 94\%\\
 \hline
\end{tabular}
\label{Comp_table}
\end{table}

\section{Discussion}

ASC$^2$End’s large-scale automation capabilities unveil a novel method to quickly perceive publicly posted information. The ASC$^2$End system provides valuable insights regarding the conformity of a candidate document corpus to the user-defined criteria. Large-scale automation of information comparison is desirable due to the possibilities of time-savings, providing more opportunities for users to perform meaningful analyses on distilled data.

As the quality, accuracy, and inference throughput of LLM-driven text analysis increases, the quality of information from our system will also increase. ASC$^2$End is widely applicable by implementing prompt engineering and RAG to obtain the desired results, removing many dependencies on pre-existing domain-specific data. Our system's methodology is intuitive and eliminates the need for LLM expertise during system operation while leveraging the user's expertise to interpret the results. The ASC$^2$End system is a new approach to applying RAG as a smart retrieval system for systematic analyses contrary to endpoints for context-driven QA tasks.

From our experimental results in Table \ref{Table5}, GPT 4 can complete a comparison assessment in 0.38 minutes demonstrating the system's scalability to much larger datasets and workflows. The use of Llama 2 70B is possible, however, it performs significantly slower and there are significant performance declines that were outlined in Table \ref{Table4}. The comparison between these two models showcases that the performance of the ASC$^2$End system is easily improvable by applying more current and capable models. Additionally, the GPT 4 model performance demonstrates that the time complexity of our system is low and is very efficient for completing large workflows with thousands of articles and documents to process, with many opportunities for improvement as LLMs become more advanced.

\begin{table}[h]
\centering
\caption{Relative run times of Llama 2 70B and GPT-4 on the comparison assessment for 1253 articles. Computational time per document is provided to interpret the effects of model runtime at scale.
}
\begin{tabular}{l c c} 
  \hline
 LLM & Min/Document & \begin{tabular}{@{}c@{}}Relative \\ Runtime\end{tabular} \\ 
  \hline Llama\,2\:70B & 1.18 & 1\\ 
 GPT 4  & \textbf{0.38} &  \textbf{0.32}\\ 
  \hline
\end{tabular}

\label{Table5}
\end{table}

ASC$^2$End eliminates redundancies of traditional human textual analysis and greatly decreases the amount of time required to extract and compare relevant information from documents. Users can quickly access relevant topics and details of documents on a larger scale, which can influence the speed at which decisions are made. This concept is especially important in the financial domain where many high-pressure decisions are time-sensitive \citep{Wegier_2015} and require accurate, publicly available information to aid in decision-making. Opening up information availability to a larger population also implies a better understanding of the underlying actions of corporations and governments that were previously "hidden in plain sight". With the application of ASC$^2$End, relevant information is properly distilled to the user, removing irrelevant and misleading information.

\subsection{\textbf{Cross-Domain Application of ASC$^2$End}}

The ASC$^2$End system was designed to function with naive generative LLMs as they can adapt to different knowledge domains by applying zero-shot prompting techniques. The prompt engineering structure used in our system can be modified to support few-shot prompting strategies for applicable knowledge domains and specific use cases where applied examples were designed. Additionally, the zero-shot prompting structure of our system can be modified to determine different focuses of analysis. Our system was designed to be robust by enabling the ability to process various data sizes through our abstractive summarization and criteria embedding modules. These modules were designed to facilitate widely reproducible results with different scopes of analysis.

\subsection{Comparison to the State-of-the-Art}

The ability for LLMs to perform information extraction (IE) has been proven to be very successful in recent advancements \citep{xu_2024}. Abstractive summarization is categorized as an IE task as it extracts and summarizes the reference text. \cite{askari_2024} demonstrated that current LLMs are not consistent on their abstraction tasks, save for Mistral 7B. \cite{askari_2024} stated that minor perturbations in the input context would greatly affect the output response and quality of the responses. However, the prompt engineering done in their study was very limited and requested single-sentence responses. As mentioned in the introduction, \cite{dixit_2023} proposed training the LLM based on labelled datasets to retain more accurate information during abstraction. They demonstrated that base models would miss certain aspects of the text that contradicted its summary. However, they stated that extensive computational and time resources would be required to perform their abstraction. Similarly, \cite{S.Liu_2023} presented the concept of semantically splitting documents instead of by token length for summarization, potentially yielding more accurate summarization results.

In our study, we performed our abstractive summarization task to shorten the input context while retaining the relevant information for analysis. We designed prompts to request specific information to be retained and direct the LLM through zero-shot prompting to extract the information of relevance. By providing more explicit instructions, minor perturbations discussed by \cite{askari_2024} have a lower impact on the summary performance. Additionally, our study aimed to create a system capable of text comparisons while being time and computationally efficient. The proposed method of \cite{dixit_2023} requires extensive fine-tuning of the LLM to generate an improved summary quality. However, the trade-off to the improved summary quality is the increased investment of computational and time resources. Our system exhibits strong information and context retention through our current prompting structure without applying model finetuning or additional context-related data splitting \citep{S.Liu_2023}.

Many SOTA methods for enhancing data along the RAG workflow have been developed to improve the implementation of RAG with LLMs \citep{gao_2024}. In our study, we classify our abstractive summarization step as a pre-retrieval process for RAG as it "rewrites" the context part of the query for our system. Additionally, we applied RAG in our system strictly as a method to inject the most relevant context to the last stage of the pipeline, in the CA module. Pre-retrieval tasks are classified under advanced RAG, thus, we can compare our RAG process to existing SOTA advanced RAG techniques.

In our study, we used abstraction to significantly shorten article lengths and preserve context, enabling similar analyses to baseline methods while minimizing resource use. \cite{zheng_2024} applies a similar concept of abstraction on the user prompt to simplify it to a "stepback" prompt such that more generic information is retrieved as context for the LLM. Applying this pre-retrieval abstraction step enables a direct search between the article and the search document, ensuring the generated response relates more to the summary than the user query. \cite{ma_2023} applies a rewriting technique that allows for direct matching of shorter text passages to the RAG system such that directly related information can be extracted.

The research proposed by \cite{ma_2023} and \cite{zheng_2024} are applied similarly to the ASC$^2$End model in which abstraction is deployed to distill the original context. However, both approaches apply much shorter passages of the original context to leverage the RAG search, functioning very similar to a QA-based task. These approaches fragment the original article and do not provide a holistic approach and analysis of the desired article. Our system applies abstraction on the whole document, a proven method from \cite{Rath_2023} to retain the complete context of the input document for the RAG process and apply it to the RAG and CA modules. Our pre-retrieval process is more complex than SOTA as we provide the distilled article instead of portions of text. ASC$^2$End demonstrates success in providing context-driven RAG outputs through our pre-retrieval pipeline. 

\subsection{\textbf{Comparison of ASC$^2$End to Related Work}}

To demonstrate the impact of the ASC$^2$End system in the field of automated text comparisons, we discuss the similarities and differences of related work. Various fields of study employ the concept of information comparison for analysis. In this section, we discuss two recent examples of information comparison methods applied to academia and sustainable finance domains.

\citet{Guo_2023}'s work automates criteria comparison for academic paper screening using LLMs. Their study presents how LLMs can elicit reasoning by comparing abstracts of scientific papers to user-defined criteria for the goal of creating a preliminary screening process for academic papers. With user-specified statements as criteria, an LLM is used to make a yes/no decision based on the abstract's conformity to the criteria. Contrary to the ASC$^2$End system, the LLM's output response lacked supporting evidence in its justification as it only outputted a yes/no binary response. In the ASC$^2$End system, the generated response provides evidence-based responses with information provided from the input context. Additionally, our system efficiently processes input text documents significantly larger than academic abstracts and can manipulate a range of criteria documents for analysis.

chatReport \citep{Ni_2023} is a financial information comparison system designed to evaluate corporate sustainability reports by using an RAG process to drive QA tasks. The goal was to compare a corporate sustainability report against the TCFD (Task Force on Climate-related Financial Disclosures) guidelines. chatReport employed a retrieval system to obtain relevant information from corporate sustainability reports that are large (i.e. typically larger than 20 pages) to run conformity and QA tasks. Their system leveraged RAG abilities for conformity assessment to a set of sustainability guidelines. However, chatReport lacked methods for automating the comparison process with the designed questions and could only analyze one candidate document at a time. In contrast, ASC$^2$End automates the comparison process and uses a RAG process as part of the workflow to only provide relevant information for the comparison assessment. Our system eliminates the dependency on multiple QA tasks to extract the information required for comparison and removes the need for user intervention to generate insights. Additionally, the ASC$^2$End system can preprocess candidate documents of any length using abstractive summarization, further demonstrating our proposed system's adaptability.

\subsection{\textbf{Limitations \& Future Work}}

ASC$^2$End acts as a proof of concept for an advanced level of information comparison in the financial domain at a large scale. However, there are limitations to the ROUGE and human-annotated metrics used to evaluate the system's performance. We also denote the current advances in the techniques used in our system and how they may be applied in future work.

Abstractive summarization is a powerful method of condensing text without losing semantic meaning, but the LLM-driven process may not accurately capture potentially important specific information. We mitigate this problem through prompt engineering. The DS module in the ASC$^2$End system improved NER performance. Specific information in the original document is often not preserved perfectly during abstraction implying that prompt changes and more powerful models may need to be applied to improve abstraction performance. Relevant information in abstractive summarization can be retained through finetuning and the future development of more compact and powerful LLMs to drive this task.

\citet{S.Jiang_2023} identified that the ROUGE metric used to evaluate the effectiveness of the generative models could be susceptible to small variations of the source data, negatively influencing the sentiment of the generated summary. This process could be manipulated by providing unhelpful inputs to prompts, negatively affecting the output. Additionally, in choosing to evaluate the abstractive summarization task with only ROUGE and not supplementing with human feedback, there was a possibility that information specificity was lost in the iterative abstractive summarization tasks. However, manually validating individual summaries and providing human feedback is not cost-effective.

The method of using RAG to retrieve relevant categories of comparison sometimes returned irrelevant information due to imbalanced attention to the provided context. This limitation is caused by how the similarity search is conducted where the criteria passages were selected based on relevancy to the target topic and \textit{not} the candidate document summary. Processing the retrieved passages could remove this unintended "noise", and should be considered for future work. Furthermore, providing automated scoring for the comparison assessment is difficult due to the complex nature of the reasoning tasks. Currently, the most accurate methods in ensuring model success during inference are to break down tasks into simple labelled QA response pairs or to employ the use of survey participants to judge the quality of the output. Using survey participants introduces response biases in terms of what is considered "good quality". Here, we limited our survey population to those with a bachelor's degree to decrease the variance in response quality. We hope that as generative LLMs become more developed, they can reliably act as their iterative feedback agent to score and improve their responses \citep{Jain_2023} without the dependency on human feedback. 

The data source used for this study presents a potential bias due to data availability constrained by the database. Factiva functions as a news archive that databases articles from a wide variety of open-source news. The ability to compile edge-case news from local news outlets, company-specific news pages, and subscription-based news sources was limited. While not the focus of this generalized solution, such data could improve robustness.

Our experimentation was constrained by hardware limitations that affected model selection and runtimes. 180B parameter models were completely out of the scope of study due to our hardware limitations. As LLMs continue to grow in complexity, additional computational resources will be needed.

\section{Conclusion}

Applications powered by generative LLMs aid in human decision-making through various emerging techniques. Here, we developed a novel system and tool, ASC$^2$End, that enables accurate, automated information comparison at scale across knowledge domains, overcoming limitations in context length and retrieval. We apply ASC$^2$End to a challenge in the financial domain, providing insights into a corpus of financial information. The system compares documents to user-defined criteria and highlights transaction-specific information and detailed analyses, resulting in time-saving efficiencies and facilitating more informed decision-making. 

In our study, we defined two different levels of task performance, machine and human level, ensuring ASC$^2$End tasks were performed by the most appropriate model. We presented the ASC$^2$End system, a powerful solution designed to perform text comparisons at scale through the applications of LLMs and surrounding technologies. The system was designed for large-scale text comparisons, minimizing information loss and addressing shortcomings in existing SOTA approaches. We found that current SOTA approaches require greater investments in time, data, and computational complexity to drive the success of their proposed methods. Additionally, existing pre-retrieval methods focus on simplifying the input for RAG instead of reducing the input size. Our system applies a novel pre-retrieval RAG pipeline and system through abstractive summarization and RAG to remove dependencies on domain-specific training data to decrease runtimes and computational complexity.

We showcased the practical uses of our system and how it can effectively implement various available models. The document summarization module results demonstrate the effectiveness of text distillation in preserving semantic meaning. These results reinforce the application of abstractive summarization as a pre-retrieval step for RAG to present relevant information and minimize loss. We found that Mistral 7B provided the best ROUGE results for summarization in shorter contexts. Survey responses indicate GPT-4 has superior reasoning and text comparison abilities compared to Llama 2-70B. Additionally, the strong survey response scoring on GPT 4's comparison assessment further reflects the effectiveness of the ASC$^2$End system in generating high-value assessment reports on financial news documents.

Our system assists in the decision-making process and can save time for analyst professionals requiring specific and semantically related information concealed in each document of a large corpus. ASC$^2$End provides an automated systematic approach to existing text comparison strategies while removing dependencies on the human operator. The application of our system in the financial world opens up the opportunity for operators to access a larger portion of data during analysis, leading to more informed decisions made by them and their organizations.

Our research provides early steps to using LLM-generated responses as a valuable aid for human-level reasoning analysis tasks. Opportunities for future development exist to improve the performance of the ASC$^2$End system further. The experimentation done in this study stands as a proof of concept to apply LLMs to automate text comparisons at scale effectively. Further hyperparameter refinement of the ASC$^2$End modules and further development of LLM models will increase the value added by the system.

\section*{Conflict of Interest Statement}

The authors declare no conflicts of interest regarding the research and findings presented in this study.
%% Run bibtex to output the sample bibliography from bib file.

%\nocite{*} - Uncomment \nocite{*} to generate bbl file or use \cite command in the text by taking the cross reference label mentioned in the bib file will generate similar NLP style bbl file. To get the exact required style modify the provided bst file

\bibliographystyle{nlplike}
\bibliography{test}

\begin{thebibliography}{}

\bibitem[Askari et~al., 2024]{askari_2024}
{\bf Askari, H.}, {\bf Chhabra, A.}, {\bf Chen, M.}, \textbf{and} {\bf Mohapatra, P.} (2024).
\newblock Assessing llms for zero-shot abstractive summarization through the lens of relevance paraphrasing.
\newblock {\em arXiv.org}.

\bibitem[Bang et~al., 2023]{Bang_2023}
{\bf Bang, Y.}, {\bf Cahyawijaya, S.}, {\bf Lee, N.}, {\bf Dai, W.}, {\bf Su, D.}, {\bf Wilie, B.}, {\bf Lovenia, H.}, {\bf Ji, Z.}, {\bf Yu, T.}, {\bf Chung, W.}, {\bf Do, Q.~V.}, {\bf Xu, Y.}, \textbf{and} {\bf Fung, P.} (2023).
\newblock A multitask, multilingual, multimodal evaluation of chatgpt on reasoning, hallucination, and interactivity.
\newblock {\em arXiv.org}.

\bibitem[Brown et~al., 2020]{Brown_2020}
{\bf Brown, T.}, {\bf Mann, B.}, {\bf Ryder, N.}, {\bf Subbiah, M.}, {\bf Kaplan, J.~D.}, {\bf Dhariwal, P.}, {\bf Neelakantan, A.}, {\bf Shyam, P.}, {\bf Sastry, G.}, {\bf Askell, A.}, {\bf Agarwal, S.}, {\bf Herbert-Voss, A.}, {\bf Krueger, G.}, {\bf Henighan, T.}, {\bf Child, R.}, {\bf Ramesh, A.}, {\bf Ziegler, D.}, {\bf Wu, J.}, {\bf Winter, C.}, {\bf Hesse, C.}, {\bf Chen, M.}, {\bf Sigler, E.}, {\bf Litwin, M.}, {\bf Gray, S.}, {\bf Chess, B.}, {\bf Clark, J.}, {\bf Berner, C.}, {\bf McCandlish, S.}, {\bf Radford, A.}, {\bf Sutskever, I.}, \textbf{and} {\bf Amodei, D.} (2020).
\newblock Language models are few-shot learners.
\newblock {\em Advances in Neural Information Processing Systems}, 33:1877--1901.

\bibitem[Bubeck et~al., 2023]{Bubeck_2023}
{\bf Bubeck, S.}, {\bf Chandrasekaran, V.}, {\bf Eldan, R.}, {\bf Gehrke, J.~A.}, {\bf Horvitz, E.}, {\bf Kamar, E.}, {\bf Lee, P.}, {\bf Lee, Y.}, {\bf Li, Y.-F.}, {\bf Lundberg, S.~M.}, {\bf Nori, H.}, {\bf Palangi, H.}, {\bf Ribeiro, M.~T.}, \textbf{and} {\bf Zhang, Y.} (2023).
\newblock Sparks of artificial general intelligence: Early experiments with gpt-4.
\newblock {\em arXiv.org}.

\bibitem[Dixit et~al., 2023]{dixit_2023}
{\bf Dixit, T.}, {\bf Wang, F.}, \textbf{and} {\bf Chen, M.} (2023).
\newblock Improving factuality of abstractive summarization without sacrificing summary quality.
\newblock {\em Association for Computational Linguistics}, pp. 902--913.

\bibitem[Espejel et~al., 2023]{Espejel_2023}
{\bf Espejel, J.~L.}, {\bf Ettifouri, E.}, {\bf Alassan, M. S.~Y.}, {\bf Chouham, E.~M.}, \textbf{and} {\bf Dahhane, W.} (2023).
\newblock Gpt-3.5, gpt-4, or bard? evaluating llms reasoning ability in zero-shot learning and performance boosting through prompts.
\newblock {\em Natural Language Processing Journal}.

\bibitem[Factiva, 2002]{factiva}
{\bf Factiva} (2002).

\bibitem[Fei et~al., 2023]{Fei_2023}
{\bf Fei, H.}, {\bf Li, B.}, {\bf Liu, Q.}, {\bf Bing, L.}, {\bf Li, F.}, \textbf{and} {\bf Chua, T.-S.} (2023).
\newblock Reasoning implicit sentiment with chain-of-thought prompting.
\newblock {\em Annual Meeting of the Association for Computational Linguistics}.

\bibitem[Frantar et~al., 2022]{Elias_2022}
{\bf Frantar, E.}, {\bf Ashkboos, S.}, {\bf Hoefler, T.}, \textbf{and} {\bf Alistarh, D.} (2022).
\newblock Gptq: Accurate post-training quantization for generative pre-trained transformers.
\newblock {\em Cornell University - arXiv}.

\bibitem[Gao et~al., 2024]{gao_2024}
{\bf Gao, Y.}, {\bf Xiong, Y.}, {\bf Gao, X.}, {\bf Jia, K.}, {\bf Pan, J.}, {\bf Bi, Y.}, {\bf Dai, Y.}, {\bf Sun, J.}, {\bf Wang, M.}, \textbf{and} {\bf Wang, H.} (2024).
\newblock Retrieval-augmented generation for large language models: A survey.
\newblock {\em arXiv.org}.

\bibitem[Gatto et~al., 2023]{Gatto_2023}
{\bf Gatto, J.}, {\bf Sharif, O.}, {\bf Seegmiller, P.}, {\bf Bohlman, P.}, \textbf{and} {\bf Preum, S.} (2023).
\newblock Text encoders lack knowledge: Leveraging generative llms for domain-specific semantic textual similarity.
\newblock {\em arXiv.org}.

\bibitem[Guo et~al., 2023]{Guo_2023}
{\bf Guo, E.}, {\bf Gupta, M.}, {\bf Deng, J.}, {\bf Park, Y.-J.}, {\bf Paget, M.}, \textbf{and} {\bf Naugler, C.} (2023).
\newblock Automated paper screening for clinical reviews using large language models.
\newblock {\em Journal of Medical Internet Research}.

\bibitem[Harrison, 2022]{langchain}
{\bf Harrison, C.} (2022).

\bibitem[Howell et~al., 2023]{Howell_2023}
{\bf Howell, K.}, {\bf Christian, G.}, {\bf Fomitchov, P.}, {\bf Kehat, G.}, {\bf Marzulla, J.}, {\bf Rolston, L.}, {\bf Tredup, J.}, {\bf Zimmerman, I.}, {\bf Selfridge, E.}, \textbf{and} {\bf Bradley, J.} (2023).
\newblock Distilled language models are economically efficient for the enterprise. ...mostly.
\newblock {\em Annual Meeting of the Association for Computational Linguistics}.

\bibitem[Huang et~al., 2023]{Huang_2023}
{\bf Huang, L.}, {\bf Yu, W.}, {\bf Ma, W.}, {\bf Zhong, W.}, {\bf Feng, Z.}, {\bf Wang, H.}, {\bf Chen, Q.}, {\bf Peng, W.}, {\bf Feng, X.}, {\bf Qin, B.}, \textbf{and} {\bf Liu, T.} (2023).
\newblock A survey on hallucination in large language models: Principles, taxonomy, challenges, and open questions.
\newblock {\em arXiv.org}.

\bibitem[HuggingFace, 2022]{openleaderboard}
{\bf HuggingFace} (2022).

\bibitem[Jain et~al., 2023]{Jain_2023}
{\bf Jain, N.}, {\bf Saifullah, K.}, {\bf Wen, Y.}, {\bf Kirchenbauer, J.}, {\bf Shu, M.}, {\bf Saha, A.}, {\bf Goldblum, M.}, {\bf Geiping, J.}, \textbf{and} {\bf Goldstein, T.} (2023).
\newblock Bring your own data! self-supervised evaluation for large language models.
\newblock {\em arXiv.org}.

\bibitem[Jaiswal and Milios, 2023]{jaiswal2023}
{\bf Jaiswal, A.} \textbf{and} {\bf Milios, E.} (2023).
\newblock Breaking the token barrier: Chunking and convolution for efficient long text classification with bert.
\newblock {\em arXiv.org}.

\bibitem[Jiang et~al., 2023a]{A.Jiang_2023}
{\bf Jiang, A.~Q.}, {\bf Sablayrolles, A.}, {\bf Mensch, A.}, {\bf Bamford, C.}, {\bf Chaplot, D.~S.}, {\bf Casas, D. D.~L.}, {\bf Bressand, F.}, {\bf Lengyel, G.}, {\bf Lample, G.}, {\bf Saulnier, L.}, {\bf Lavaud, L.~R.}, {\bf Lachaux, M.-A.}, {\bf Stock, P.}, {\bf Scao, T.~L.}, {\bf Lavril, T.}, {\bf Wang, T.}, {\bf Lacroix, T.}, \textbf{and} {\bf Sayed, W.~E.} (2023)a.
\newblock Mistral 7b.
\newblock {\em arXiv.org}.

\bibitem[Jiang et~al., 2023b]{S.Jiang_2023}
{\bf Jiang, S.}, {\bf Kadhe, S.}, {\bf Zhou, Y.}, {\bf Cai, L.}, \textbf{and} {\bf Baracaldo, N.} (2023)b.
\newblock Forcing generative models to degenerate ones: The power of data poisoning attacks.
\newblock {\em null}.

\bibitem[Johnson et~al., 2017]{Johnson_2017}
{\bf Johnson, J.~R.}, {\bf Johnson, J.}, {\bf Douze, M.}, \textbf{and} {\bf Jégou, H.} (2017).
\newblock Billion-scale similarity search with gpus.
\newblock {\em IEEE Transactions on Big Data}.

\bibitem[Johnson-Laird, 2010]{Johnson-Laird_2010}
{\bf Johnson-Laird, P.~N.} (2010).
\newblock Inaugural article by a recently elected academy member mental models and human reasoning.
\newblock {\em Proceedings of the National Academy of Sciences of the United States of America}.

\bibitem[Kojima et~al., 2022]{Kojima_2022}
{\bf Kojima, T.}, {\bf Gu, S.~S.}, {\bf Reid, M.}, {\bf Matsuo, Y.}, \textbf{and} {\bf Iwasawa, Y.} (2022).
\newblock Large language models are zero-shot reasoners.
\newblock 35:22199--22213.

\bibitem[Kumar et~al., 2023]{Kumar_2023}
{\bf Kumar, J.}, {\bf Shekhar, S.}, \textbf{and} {\bf Gupta, R.} (2023).
\newblock Hindi news article's headline generation based on abstractive text summarization.
\newblock {\em 2023 International Conference on Network, Multimedia and Information Technology (NMITCON)}.

\bibitem[Li et~al., 2023]{Li_2023}
{\bf Li, X.}, {\bf Zhu, X.}, {\bf Ma, Z.}, {\bf Liu, X.}, \textbf{and} {\bf Shah, S.} (2023).
\newblock Are chatgpt and gpt-4 general-purpose solvers for financial text analytics? an examination on several typical tasks. arxiv.

\bibitem[Lin, 2004]{Lin_2004}
{\bf Lin, C.-Y.} (2004).
\newblock Rouge : a package for automatic evaluation of summaries.
\newblock {\em Proceedings of the Workshop on Text Summarization Branches Out, 2004}.

\bibitem[Liu et~al., 2023a]{H.Liu_2023}
{\bf Liu, H.}, {\bf Ning, R.}, {\bf Teng, Z.}, {\bf Liu, J.}, {\bf Zhou, Q.}, \textbf{and} {\bf Zhang, Y.} (2023)a.
\newblock Evaluating the logical reasoning ability of chatgpt and gpt-4.
\newblock {\em arXiv.org}.

\bibitem[Liu et~al., 2023b]{N.Liu_2023}
{\bf Liu, N.~F.}, {\bf Lin, K.}, {\bf Hewitt, J.}, {\bf Paranjape, A.}, {\bf Bevilacqua, M.}, {\bf Petroni, F.}, \textbf{and} {\bf Liang, P.} (2023)b.
\newblock Lost in the middle: How language models use long contexts.
\newblock {\em arXiv.org}.

\bibitem[Liu and Healey, 2023]{S.Liu_2023}
{\bf Liu, S.} \textbf{and} {\bf Healey, C.~G.} (2023).
\newblock Abstractive summarization of large document collections using gpt.
\newblock {\em arXiv.org}.

\bibitem[Lu et~al., 2021]{Lu_2021}
{\bf Lu, Y.}, {\bf Bartolo, M.}, {\bf Moore, A.}, {\bf Riedel, S.}, \textbf{and} {\bf Stenetorp, P.} (2021).
\newblock Fantastically ordered prompts and where to find them: Overcoming few-shot prompt order sensitivity.
\newblock {\em Annual Meeting of the Association for Computational Linguistics}.

\bibitem[Ma et~al., 2023]{ma_2023}
{\bf Ma, X.}, {\bf Gong, Y.}, {\bf He, P.}, {\bf Zhao, H.}, \textbf{and} {\bf Duan, N.} (2023).
\newblock Query rewriting for retrieval-augmented large language models.
\newblock {\em arXiv.org}.

\bibitem[Majumder et~al., 2016]{Majumder_2016}
{\bf Majumder, G.}, {\bf Pakray, P.}, {\bf Pakray, P.}, {\bf Gelbukh, A.}, \textbf{and} {\bf Pinto, D.} (2016).
\newblock Semantic textual similarity methods, tools, and applications: A survey.
\newblock {\em Computación Y Sistemas}.

\bibitem[Muennighoff et~al., 2022]{Muennighoff_2022}
{\bf Muennighoff, N.}, {\bf Tazi, N.}, {\bf Magne, L.}, \textbf{and} {\bf Reimers, N.} (2022).
\newblock Mteb: Massive text embedding benchmark.
\newblock {\em Conference of the European Chapter of the Association for Computational Linguistics}.

\bibitem[Navarro et~al., 2022]{Navarro_2022}
{\bf Navarro, D.~F.}, {\bf Dras, M.}, \textbf{and} {\bf Berkovsky, S.} (2022).
\newblock Few-shot fine-tuning sota summarization models for medical dialogues.
\newblock {\em North American Chapter of the Association for Computational Linguistics}.

\bibitem[Ni et~al., 2023]{Ni_2023}
{\bf Ni, J.}, {\bf Bingler, J.}, {\bf Colesanti-Senni, C.}, {\bf Kraus, M.}, {\bf Gostlow, G.}, {\bf Schimanski, T.}, {\bf Stammbach, D.}, {\bf Vaghefi, S.}, {\bf Wang, Q.}, {\bf Webersinke, N.}, {\bf Wekhof, T.}, {\bf Yu, T.}, \textbf{and} {\bf Leippold, M.} (2023).
\newblock Chatreport: Democratizing sustainability disclosure analysis through llm-based tools.
\newblock {\em Conference on Empirical Methods in Natural Language Processing}.

\bibitem[OpenAI, 2023a]{OpenAI_2023}
{\bf OpenAI} (2023)a.
\newblock Gpt-4 technical report.
\newblock {\em arXiv.org}.

\bibitem[OpenAI, 2023b]{tokenizer}
{\bf OpenAI} (2023)b.
\newblock Tokenizer.
\newblock Accessed on December 5, 2023.

\bibitem[Pan et~al., 2022]{Pan_2022}
{\bf Pan, F.}, {\bf Canim, M.}, {\bf Glass, M.~R.}, {\bf Gliozzo, A.}, \textbf{and} {\bf Hendler, J.} (2022).
\newblock End-to-end table question answering via retrieval-augmented generation.
\newblock {\em arXiv.org}.

\bibitem[Radford et~al., 2019]{Radford_2019}
{\bf Radford, A.}, {\bf Wu, J.}, {\bf Child, R.}, {\bf Luan, D.}, {\bf Amodei, D.}, \textbf{and} {\bf Sutskever, I.} (2019).
\newblock Language models are unsupervised multitask learners.

\bibitem[Ram et~al., 2023]{Ram_2023}
{\bf Ram, O.}, {\bf Levine, Y.}, {\bf Dalmedigos, I.}, {\bf Muhlgay, D.}, {\bf Shashua, A.}, {\bf Leyton-Brown, K.}, \textbf{and} {\bf Shoham, Y.} (2023).
\newblock In-context retrieval-augmented language models.
\newblock {\em Transactions of the Association for Computational Linguistics}.

\bibitem[Rath et~al., 2023]{Rath_2023}
{\bf Rath, M.}, {\bf Banerjee, S.}, \textbf{and} {\bf Swain, T.} (2023).
\newblock Fine tuning auto regressive llms for long document abstractive summarization.
\newblock {\em 2023 IEEE 2nd International Conference on Industrial Electronics: Developments \& Applications (ICIDeA)}.

\bibitem[Shapira et~al., 2018]{Shapira_2018}
{\bf Shapira, O.}, {\bf Gabay, D.}, {\bf Ronen, H.}, {\bf Bar-Ilan, J.}, {\bf Amsterdamer, Y.}, {\bf Nenkova, A.}, \textbf{and} {\bf Dagan, I.} (2018).
\newblock Evaluating multiple system summary lengths: A case study.
\newblock In {\bf Riloff, E.}, {\bf Chiang, D.}, {\bf Hockenmaier, J.}, \textbf{and} {\bf Tsujii, J.}, editors, {\em Proceedings of the 2018 Conference on Empirical Methods in Natural Language Processing}, pp. 774--778, Brussels, Belgium. Association for Computational Linguistics.

\bibitem[Slimani, 2013]{Slimani_2013}
{\bf Slimani, T.} (2013).
\newblock Description and evaluation of semantic similarity measures approaches.
\newblock {\em International Journal of Computer Applications}.

\bibitem[Sun et~al., 2023]{sun2023}
{\bf Sun, X.}, {\bf Dong, L.}, {\bf Li, X.}, {\bf Wan, Z.}, {\bf Wang, S.}, {\bf Zhang, T.}, {\bf Li, J.}, {\bf Cheng, F.}, {\bf Lyu, L.}, {\bf Wu, F.}, \textbf{and} {\bf Wang, G.} (2023).
\newblock Pushing the limits of chatgpt on nlp tasks.
\newblock {\em arXiv.org}.

\bibitem[Touvron et~al., 2023]{Touvron_2023}
{\bf Touvron, H.}, {\bf Martin, L.}, {\bf Stone, K.~R.}, {\bf Albert, P.}, {\bf Almahairi, A.}, {\bf Babaei, Y.}, {\bf Bashlykov, N.}, {\bf Batra, S.}, {\bf Bhargava, P.}, {\bf Bhosale, S.}, {\bf Bikel, D.~M.}, {\bf Blecher, L.}, {\bf Ferrer, C.~C.}, {\bf Chen, M.}, {\bf Cucurull, G.}, {\bf Esiobu, D.}, {\bf Fernandes, J.}, {\bf Fu, J.}, {\bf Fu, W.}, {\bf Fuller, B.}, {\bf Gao, C.}, {\bf Goswami, V.}, {\bf Goyal, N.}, {\bf Hartshorn, A.}, {\bf Hosseini, S.}, {\bf Hou, R.}, {\bf Inan, H.}, {\bf Kardas, M.}, {\bf Kerkez, V.}, {\bf Khabsa, M.}, {\bf Kloumann, I.~M.}, {\bf Korenev, A.}, {\bf Koura, P.~S.}, {\bf Lachaux, M.-A.}, {\bf Lavril, T.}, {\bf Lee, J.}, {\bf Liskovich, D.}, {\bf Lu, Y.}, {\bf Mao, Y.}, {\bf Martinet, X.}, {\bf Mihaylov, T.}, {\bf Mishra, P.}, {\bf Molybog, I.}, {\bf Nie, Y.}, {\bf Poulton, A.}, {\bf Reizenstein, J.}, {\bf Rungta, R.}, {\bf Saladi, K.}, {\bf Schelten, A.}, {\bf Silva, R.}, {\bf Smith, E.~M.}, {\bf Subramanian, R.}, {\bf Tan, X.}, {\bf Tang, B.}, {\bf Taylor, R.}, {\bf
  Williams, A.}, {\bf Kuan, J.~X.}, {\bf Xu, P.}, {\bf Yan, Z.}, {\bf Zarov, I.}, {\bf Zhang, Y.}, {\bf Fan, A.}, {\bf Kambadur, M.}, {\bf Narang, S.}, {\bf Rodriguez, A.}, {\bf Stojnic, R.}, {\bf Edunov, S.}, \textbf{and} {\bf Scialom, T.} (2023).
\newblock Llama 2: Open foundation and fine-tuned chat models.
\newblock {\em arXiv.org}.

\bibitem[Trivedi et~al., 2022]{Trivedi_2022}
{\bf Trivedi, H.}, {\bf Balasubramanian, N.}, {\bf Khot, T.}, \textbf{and} {\bf Sabharwal, A.} (2022).
\newblock Interleaving retrieval with chain-of-thought reasoning for knowledge-intensive multi-step questions.
\newblock {\em Annual Meeting of the Association for Computational Linguistics}.

\bibitem[Wegier et~al., 2015]{Wegier_2015}
{\bf Wegier, P.}, {\bf Wegier, P.}, {\bf Wegier, P.}, \textbf{and} {\bf Spaniol, J.} (2015).
\newblock The effect of time pressure on risky financial decisions from description and decisions from experience.
\newblock {\em PLOS ONE}.

\bibitem[Wei et~al., 2022]{Wei_2022}
{\bf Wei, J.}, {\bf Wang, X.}, {\bf Schuurmans, D.}, {\bf Bosma, M.}, {\bf ichter, b.}, {\bf Xia, F.}, {\bf Chi, E.}, {\bf Le, Q.~V.}, \textbf{and} {\bf Zhou, D.} (2022).
\newblock Chain-of-thought prompting elicits reasoning in large language models.
\newblock 35:24824--24837.

\bibitem[Xiao et~al., 2023]{Xiao_2023}
{\bf Xiao, S.}, {\bf Liu, Z.}, {\bf Zhang, P.}, \textbf{and} {\bf Muennighoff, N.} (2023).
\newblock C-pack: Packaged resources to advance general chinese embedding.
\newblock {\em arXiv.org}.

\bibitem[Xu et~al., 2024]{xu_2024}
{\bf Xu, D.}, {\bf Chen, W.}, {\bf Peng, W.}, {\bf Zhang, C.}, {\bf Xu, T.}, {\bf Zhao, X.}, {\bf Wu, X.}, {\bf Zheng, Y.}, {\bf Wang, Y.}, \textbf{and} {\bf Chen, E.} (2024).
\newblock Large language models for generative information extraction: A survey.
\newblock {\em arXiv.org}.

\bibitem[Zheng et~al., 2024]{zheng_2024}
{\bf Zheng, H.~S.}, {\bf Mishra, S.}, {\bf Chen, X.}, {\bf Cheng, H.-T.}, {\bf Chi, E.~H.}, {\bf Le, Q.~V.}, \textbf{and} {\bf Zhou, D.} (2024).
\newblock Take a step back: Evoking reasoning via abstraction in large language models.
\newblock {\em arXiv.org}.

\end{thebibliography}

\label{lastpage}

\end{document}